\documentclass[a4paper,twoside]{article}

\usepackage{epsfig}
\usepackage{subcaption}
\usepackage{calc}
\usepackage{amssymb}
\usepackage{amstext}
\usepackage{amsmath}
\usepackage{amsthm}
\usepackage{multicol}
\usepackage{pslatex}
\usepackage{natbib}
\let\cite\citep
\usepackage{graphicx}
\usepackage{booktabs}
\usepackage{algorithm2e}
\usepackage[bottom]{footmisc}
\usepackage[hyphens]{url}
\usepackage{SCITEPRESS} 

\begin{document}
\title{CLASH: Detecting Contradictions in Multi-Document Policy Corpora
with a Two-Stage LLM framework}

\author{\authorname{Abdullah As Sami, Youjian Tang, Trung Lam and John Licato}
\affiliation{Advancing Machine and Human Reasoning (AMHR) Lab\\
Bellini College of Artificial Intelligence, Cybersecurity and Computing\\University of South Florida}}

\abstract{
Different individuals revise multiple organizational documents at different times. As a result, rules that apply to the same employee may contradict each other, and these conflicts often go unnoticed until someone follows one rule while violating another. Manual review is not scalable. A corpus of 1,700 clauses yields over 1.5 million possible pairs. We present CLASH, a framework based on large language models (LLM) that operates in two stages. First, it retrieves potentially conflicting clause pairs using context-enriched embeddings and contradiction-aware reranking, and then verifies each remaining pair within its source document. On two multi-document policy corpora and a contract corpus, CLASH achieves F1 scores between .840 and .921, while natural language inference (NLI) and whole-document prompting reach at most .340 in matched comparisons. The gain comes from giving the model document context at both stages. When applied unchanged to public corpora, the framework identified five actual contradictions, three in a university's regulations and two in a federal agency's personnel policies. We manually verified each against the source documents.}

\onecolumn \maketitle \normalsize \setcounter{footnote}{0} \vfill

\begin{figure*}[htb!]
    \centering
    \includegraphics[width=\textwidth]{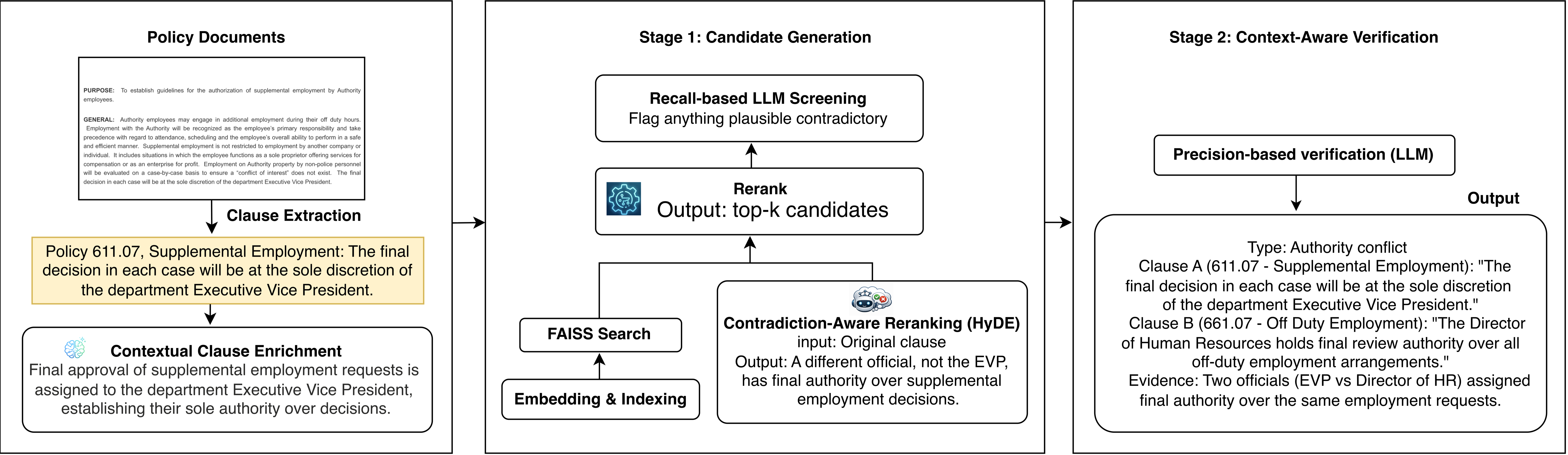}
    \caption{The CLASH framework. Contextual clause enrichment improves retrieval recall. Stage 1 enriches each clause with local context. It then retrieves and reranks candidate pairs toward contradiction. Stage 2 re-reads each surviving pair.}
    \label{fig:pipeline}
\end{figure*}

\section{\uppercase{Introduction}}\label{sec:intro}

Organizations use written policies to guide their employees. An HR handbook says how leave is approved, a university regulation says who may file a grievance, a contract says what must stay confidential. These policies undergo revisions by different authors over the years, and those revisions may not preserve the coherence of the entire set of policies. When one policy is updated and a related one is not, the two can end up giving employees incompatible instructions. One clause requires supervisor approval for a request while another assigns that decision to a different office, or two policies state two different deadlines for the same form. Such contradictions can lead an employee to follow one policy and unknowingly violate another, exposing the organization's compliance risk and legal exposure. In practice, organizations rely mainly on manual review to verify consistency across their policy documents. Some recent large language model (LLM) based systems have begun to automate this. Recently, the software industry has emerged to solve contradiction detection for organizations by providing solutions as Software as a Service (SaaS) without any established benchmark. A recent work used a synthetic legal-style dataset to validate a contradiction framework \citep{mantravadi2025legalwiz}. However, results from generated data might not apply to real policy documents. Moreover, contradiction detection has traditionally been formulated as a pairwise classification problem through natural language inference (NLI) \citep{maccartney2009natural}. But NLI-based contradiction detection methods are not directly applicable in multi document policy corpora as the sentence pair is not known. Furthermore, identifying contradictions in organizational policy documents is fundamentally different from sentence-pair classification and has attracted little attention because it involves four bottlenecks. 

First, there is the issue of scale. The number of possible clause pairs grows very quickly as the collection gets bigger, reaching millions of pairs even for small corpora. Second, contradictions in policy are rarely obvious. Two clauses may clash over who holds authority, which number applies, or which procedure governs, without ever using opposing words. Third, context sensitivity—a clause often cannot be judged without its governing document, its target population, and the exceptions sitting next to it. Finally, searching the whole document at once is not a robust strategy for this task. Accuracy drops as important details get buried deeper in long inputs~\citep{liu2024lost}, and the effect is especially strong for tasks that require careful comparison of two passages that could be anywhere in the corpora. Even when a corpus fits in a single prompt, contradiction discovery still requires comparing every clause with every other clause, and an LLM given the full text has no mechanism for working through those $O(n^2)$ comparisons one by one. 

These constraints motivate a pairwise clause-level formulation of contradiction detection. We present CLASH, a two-stage framework designed to detect contradiction across multi document corpora. Figure~\ref{fig:pipeline} shows the framework end-to-end, from a raw policy corpus to a verified contradiction. Stage~1 (candidate generation) casts a wide net over the pair space, retrieving plausible contradiction candidates and screening them with a recall-based LLM that keeps anything plausibly contradictory. Stage~2 is the verification stage that uses each candidate's original document context and provides a verdict. This decomposition makes contradiction detection tractable in large policy collections, allowing the verifier to reason over subtle, context-dependent conflicts. 

This framing motivates three research questions:

\begin{description}
    \item[RQ1.] Does dividing contradiction detection into candidate generation and pairwise verification outperform existing NLI-based classification and whole-document prompting on multi-document policy corpora?
    \item[RQ2.] Does enriching clauses with document context enhance candidate retrieval, and does including surrounding clauses, agentic tool use or type-specific skills improve verification performance?
    \item[RQ3.] Can the framework reveal genuine, previously unknown contradictions in real policy corpora?
\end{description}

\noindent
We make the following contributions:
\begin{enumerate}
    \item We present CLASH, a two-stage framework for contradiction detection in organizational multi-document policy corpora. To the best of our knowledge, no previous work detects contradictions in this setting. Existing approaches classify pairs that are already given or mine injected contradictions from synthetic corpora. Our proposed method outperforms both NLI and whole-document prompting by 23--51 F1 points across three actual policy corpora.
    \item We show that contextual clause representation increases retrieval recall from 74\% to 98\% and using the context of the document reduces false positives from 29 to 5 for LLM-based verification. This single pass verification also exceeds agentic tool-based verification while type-specific skills only help recall for small open-source LLMs in verification.
    \item We detect five real contradictions in the original policies of a university and a federal civilian personnel manual. Each candidate was flagged by multiple models and then manually verified.
 \end{enumerate}

\section{\uppercase{Related Work}}\label{sec:related}

Contradiction detection has traditionally been studied as textual inference, in which a model predicts whether a premise--hypothesis pair is contradictory \citep{bowman2015snli,williams2018multinli}. Later work brings this formulation to legal and regulatory text, from norm-conflict detection \citep{aires2017norm,huang2024neuralconflict} to a prompt-based LLM classification \citep{schumann2026detection}. All of these assume the relevant pair is given; cross-document variants likewise assumes a known mapping. For example, \citet{sai2023detecting} check internal process documents against the regulations they implement, \citet{abualhaija2024ai} track how legal changes propagate to dependent documents, and \citet{he2026foundations} verify consistency over curated atomic facts. A policy corpus does not provide predefined pairs, known mappings, or a clear list of statements. A method must therefore identify which clauses to compare. Whole-document prompting avoids the pairing step, but models lose information in lengthy inputs \citep{liu2024lost}, and confidently pick one side of a conflict rather than flagging it \citep{kurfali2025conflicting}. RQ1 tests whether separating pair discovery from verification outperforms an NLI classifier and whole-document prompting.

Once a candidate pair is found, the verifier's judgement depends on context. ContractNLI shows that contradiction judgments in contracts depend on surrounding context, exceptions, and clause scope, and that evidence spans matter \citep{koreeda2021contractnli}. \citet{gokul2025ragcontradiction} find that LLM validation performance varies widely across contradiction types and prompting strategies.  However, it remains unclear whether context can improve two-stage contradiction detection in a multi-document setting. RQ2 examines these two functions independently. For retrieval, we prepend a document summary to each clause; for verification, we give the verifier the surrounding clauses of each candidate pair.

Another approach to improving verification is to add capability. ALICE combines formal logic with LLM reasoning to detect contradictions in software requirements and finds that explicit reasoning constraints outperform unconstrained prompting \citep{gaertner2024alice}. Tool-MAD has several agents to debate a claim using retrieval and web-search \citep{jeong2026toolmad}. \citet{ma2026gkmad} guide such debates with external knowledge. A survey covers procedural skills added to agents at inference \citep{jiang2026sok}. All of these are designed for situations where the model must find or create its own evidence, yet none asks whether added capability is still useful where the relevant context is simply provided. Context and capability are thus competing explanations for reliable verification. We test this in RQ2 by equipping the verifier with retrieval tools, multi-round agent loops, and type-specific skill, and comparing each against a verifier that simply reads the surrounding clauses. This tests whether reliability comes from restored context or added capability.

Most contradiction datasets are created by adding conflicts to documents or by labeling pairs already identified by human. A review of 46 regulatory-conflict studies found that 87\% of conflict cases were synthetically generated, raising questions about their similarity to real-world conflicts \citep{schumann2024literature}. ContraDoc adds contradictions to real documents and finds that current LLMs often miss them in longer texts \citep{li2024contradoc}. CLAUSE uses LLMs to introduce legal and internal contradictions into more than 7,500 real contracts, showing that models miss many subtle cases \citep{choudhury2026better}. \citet{schumann2025contradictions} gather contradictory regulatory sentences pairs and validate their annotation rules by interviewing internal auditors, who confirm that such conflicts arise in manual document comparisons. \citet{mantravadi2025legalwiz} combine mining and verification, but test on a synthetic dataset. \citet{kozlova2026contradiction} match Russian court rulings to their known statutes in advance, and note that no effective method exists for finding contradictions between a document and an entire repository. None of these studies shows that a method can find real, previously unknown contradictions. We address this in RQ3 by auditing CLASH's detection against the source policies of a public university and a federal civilian personnel manual, confirming 5 contradictions that already existed in those policies.

\section{\uppercase{Methodology}}\label{sec:method}

\subsection{Task Definition}\label{sec:formulation}

Let $\mathcal{D} = \{D_1, \ldots, D_N\}$ be a corpus of $N$ policy documents. Each document $D_i$ is segmented into clauses $\mathcal{C}_i =
  \{c_{i,1}, \ldots, c_{i,n_i}\}$, and the full clause set is $\mathcal{C} =
  \bigcup_{i=1}^{N} \mathcal{C}_i$ with $|\mathcal{C}| = M$.

Two clauses $c_a \in \mathcal{C}_i$ and $c_b \in \mathcal{C}_j$ form a contradiction if some situation falls within the scope of both clauses and no single course of action can satisfy the requirements of both. Both conditions are essential in practical scenarios. A leave policy for faculty and a stricter policy for temporary staff assign different rules but never apply to the same employee, so they do not conflict. In contrast, two clauses that assign final approval authority over the same request to different officials do conflict. A contradiction is a self-contradiction when $i=j$ (i.e., the document contradicts itself) and pairwise when $i \neq j$. The objective is to identify all contradiction pairs in $\mathcal{C}$. 

\subsection{Corpus Preparation and Clause Extraction}\label{sec:corpus}

We apply the framework to four corpora, including two collections of human resources (HR) policies, one from a local government entity (GOV) and one from University of South Florida HR Regulations (USF), as well as a contract corpus (NDA), and a set of federal civilian personnel policies (CPM), which is used as a case study. Section~\ref{sec:datasets} provides detailed descriptions of all four corpora. We convert each policy document into JSON, then extracted clauses by segmenting the body text at sentence boundaries. Clauses shorter than 15 characters are discarded, as segments below this length are almost always headings, numbering fragments, or parsing artifacts.

\subsection{Contextual Clause Enrichment}\label{sec:contextual}

When a clause is taken out of its document, it loses much of its meaning. For example, ``The supervisor shall approve all requests'' could refer to leave, overtime, or procurement. The sentence does not specify which situation it covers or who it applies to. Inspired by Anthropic's contextual retrieval approach \citep{anthropic2024contextual} and recent evidence that document-level context improves chunk-level retrieval embeddings \citep{conti2025context}, we add a one-sentence summary before each clause to explain what it does and who it applies to. To create the summary for clause $c_{i,k}$, the LLM uses the summary of document $D_i$ (generated once for each document), the two clauses before and after, and the clause itself, and then produces a single sentence. We fixed this window at two clauses per side without tuning it, since neighboring clauses usually carry the disambiguating conditions, while a wider window mixes in unrelated rules.

The contextual representation is

\begin{equation}
    \tilde{c}_{i,k} = x_{i,k} \oplus c_{i,k},
\end{equation}

where $x_{i,k}$ is the contextual sentence generated for clause $c_{i,k}$ and $\oplus$ denotes string concatenation. In the example above, the LLM generates the contextual sentence ``Establishes supervisor approval requirement for vacation leave requests,'' and adds it before the clause for embedding. Stage 1 then embeds and retrieves using these enriched representations $\tilde{c}_{i,k}$. The added sentence brings together clauses that are topically related but use different wording, making it easier to find contradictions across policies~\citep{conti2025context, jalocha2026crawler}.

\subsection{Stage 1: Candidate Generation}\label{sec:stage1_method}

Stage~1 aims to reduce the pairwise search space $\binom{M}{2}$ to a manageable set of plausible contradiction candidates. It consists of three components: embedding and indexing, contradiction-aware re-ranking, and recall-based LLM screening.

\subsubsection{Embedding and Indexing}

We embed each enriched representation $\tilde{c}_{i,k}$ with Nomic Embed~\citep{nussbaum2024nomic}. This open-weight embedding model maps text to a 768-dimensional vector so semantically similar clauses lie close together. We chose it because it runs locally at no API cost and performs competitively on retrieval benchmarks. Next, we index these vectors with FAISS \citep{douze2025faisslibrary}, a library for fast similarity search over dense vectors. Each clause retrieves its top-$K$ neighbors by cosine similarity. Sometimes, clause $a$ retrieves $b$, and $b$ also retrieves $a$. To address this, we remove duplicate pairs. This leaves about 31,000 unique candidates for the GOV corpus ($K = 25$, $M = 1{,}755$).

\subsubsection{Contradiction-Aware Reranking (HyDE)}\label{sec:hyde}

Bi-encoder retrieval finds semantically similar clauses, but a contradicting clause is only topically similar while being logically opposed, which often leaves it just outside the top-$K$. Adapting the HyDE approach~\citep{gao2023hyde}, we prompt an LLM to write, for each clause $c$, a hypothetical contradicting statement $h(c)$: a plausible clause that would contradict $c$ if it existed in the corpus. For the clause ``Employees are \textbf{not permitted} to schedule leave during their final two
weeks,'' the hypothesis is ``Employees are \textbf{permitted} to schedule leave during their final two weeks.'' Because $h(c)$ shares the logical polarity of the target, its embedding lands closer to the real contradicting clause than the embedding of $c$ itself. We use $h(c)$ as a re-ranking signal rather than a replacement query. FAISS first retrieves the top-50 candidates by $\cos(e_c, e_x)$, where $e_\cdot$ denotes a clause's embedding. This number is twice the final retrieval depth, allowing the reranker to promote pairs that would otherwise fall outside the top 25 based solely on similarity. Each candidate $x$ is then rescored as

\begin{equation}
\text{score}(x) = (1 - \alpha)\cos(e_c, e_x) + \alpha\cos(e_{h(c)}, e_x),
\end{equation}

and the top-25 are retained. We explain the choice of $K$ in Section~\ref{sec:retrieval_ablations}. The parameter $\alpha$ balances semantic similarity with contradiction-awareness. We set $\alpha = 0.5$, weighting both signals equally. A post-hoc sweep over 0.25, 0.5, and 0.75 changed recall at the $K = 25$ operating point by at most one ground-truth pair and Mean Reciprocal Rank (MRR) by under 0.01. So, the results are not sensitive to this choice.

\subsubsection{Recall-based LLM Screening}\label{sec:stage1} We use GPT-5 mini to screen retrieved candidate pairs in batches and flag any pair showing a plausible contradiction \citep{openai_gpt5_mini}. Screening must process tens of thousands of pairs, and should keep plausible candidates. A small, fast model is sufficient to keep this stage's cost low. We made the screening permissive by keeping uncertain pairs because Stage~2 verification will filter false positives (FP).

\subsection{Stage 2: Context-Aware Verification}\label{sec:stage2}

Stage~1 produces hundreds of false postives as it achieves high recall (92--100\%) but low precision (9--36\%). So, we re-evaluate each Stage~1 candidate individually to filter false positives.

Policies often state a general rule in one clause and their exception in the next. Without surrounding context, the general rule appears to contradict a clause elsewhere in the corpus. With adjacent text visible, the exception resolves the conflict. That's why for each clause in a flagged pair, we retrieve $\pm 5$ surrounding clauses from its source document. We justify the choice of $N = 5$ with a window sweep in Section~\ref{sec:ablation_ctx_window}. For self-contradictions, both clauses come from the same policy. For pairwise contradictions, an important subtlety arises. A clause may be retrieved from a host document even if it originates in a different policy. We determine the true source from Stage~1 metadata and pull context from the source policy instead of the host. The target clause is located in its source document via word-overlap matching ($\geq 0.25$ threshold). This threshold is set low enough to tolerate small text differences from extraction, since within a single document the true clause is the obvious best match. Then, the matched clause is marked with a \texttt{>>>} prefix so that the LLM knows which clause in the context window is under evaluation.

\section{\uppercase{Experimental Setup}}\label{sec:setup}

\subsection{Hypotheses}\label{sec:hypotheses}
We derive five hypotheses from our research questions, each with its deciding metric and the section that tests it.

\noindent\textbf{H1} (RQ1). Separating candidate generation from pairwise verification improves contradiction-detection performance over NLI-based classification and whole-document prompting. Evaluated by F1 (Sec.~\ref{sec:baseline_results}).

\noindent\textbf{H2a} (RQ2). Contextual clause representation improves candidate-retrieval recall. Evaluated by Recall@20 and MRR (Sec.~\ref{sec:ctx_vs_raw}).

\noindent\textbf{H2b} (RQ2). Providing surrounding clauses during verification improves verification performance. Evaluated by F1 and false positives (Sec.~\ref{sec:stage2_results}).

\noindent\textbf{H2c} (RQ2). Tool-using agentic verification or using skills outperforms a fixed context window. Evaluated by verification F1 (Sec.~\ref{sec:skills}).

\noindent\textbf{H3} (RQ3). Some detections matching no ground-truth entry are genuine conflicts present in the source documents. Evaluated by manual review against the original documents (Sec.~\ref{sec:organic}).
  
\subsection{Datasets}\label{sec:datasets}

We evaluate our approach on three benchmark corpora with annotated ground truth and apply the framework to a fourth public corpus as a case study.

\paragraph{Local Government Entity (GOV).}
This corpus contains 100 HR policy documents from a local government entity (name withheld), covering compensation, benefits, hiring, performance evaluation, safety, and IT. Among these, 71 are core HR policies and 29 are distractor documents included to test false-positive resistance. The ground truth contains 100 annotated contradiction pairs (51 self-contradictions, 49 pairwise) across seven contradiction types, with subtlety ratings. The corpus is not publicly available.

\paragraph{University of South Florida HR Regulations (USF).}
This corpus includes 43 university HR regulations and policies\footnote{Available at \url{https://www.usf.edu/hr/policies-and-procedures/policies.aspx}, accessed 2026-04-09.} covering faculty governance, staff benefits, hiring, and disciplinary procedures. Ground truth contains 50 annotated contradiction pairs (27 self, 23 pairwise). This corpus features parallel policies for different employee groups (faculty, administration staff, temporary employees), creating natural scope-distinction challenges.

\paragraph{Non-Disclosure Agreements (NDA).}
This corpus includes 52 NDAs from the ContractNLI dataset \citep{koreeda2021contractnli}, evaluated against 17 legal hypotheses. Ground truth contains 409 annotated contradiction pairs. This corpus tests the framework's generalization beyond HR policy to contract analysis.

For the NDA corpus, we use the ContractNLI dataset~\citep{koreeda2021contractnli}. This dataset contains 607 NDAs, each annotated with 17 legal hypotheses. We sampled 52 NDAs from its training, development, and test splits covering the scope of confidentiality, permitted disclosures, obligations upon termination, and related topics.

\paragraph{Civilian Personnel Management (CPM).}
We collected 27 volumes of DoD Instruction 1400.25, titled Civilian Personnel Management,\footnote{Available at \url{https://www.esd.whs.mil/DD/DoD-Issuances/140025/}, accessed 2026-07-29.} to test the framework on new documents. These volumes are a subset of the full 1400.25 series. From 750 pages covering pay, leave, benefits, retirement, injury compensation, and position classification, we extracted 2{,}449 clauses. This corpus is not included in any benchmark table because it has no ground truth. We ran the framework, filtered confirmed detections using multi-model consensus, and manually checked the remaining results against the source documents. 

\subsection{Ground Truth Construction}\label{sec:gt}

Real policy corpora do not include labeled contradictions. We built the GOV and USF ground truth by pairing each corpus clause with a conflicting clause we wrote. For each entry, we select a clause that states an enforceable rule. We draft its conflicting counterpart under a fixed quota over the seven contradiction types (Table~\ref{tab:taxonomy}) and decide its placement. Each type has its own transformation: we inverted obligations (negation), changed required values (numerical), shifted deadlines (temporal), reassigned final approval (authority), prescribed a different procedure for the same action (process), added a specific provision that conflicts with the general rule (specificity), and permitted what the other clause prohibits (permission/prohibition). A self-contradiction places the new clause in the same document. A pairwise contradiction places it in a different policy and rewrites it in the host policy's vocabulary, so the two clauses do not share the same words. Each pair is rated obvious or subtle. Obvious pairs conflict through explicit negation or directly incompatible values, and subtle pairs require reasoning about scope, authority, or implied obligations. The result is 100 GOV pairs (51 self, 49 pairwise; 52 obvious, 48 subtle) and 50 USF pairs (27 self, 23 pairwise; 27 obvious, 23 subtle).

ContractNLI annotates each of its 607 NDAs against 17 fixed legal hypotheses, labeling every document--hypothesis combination as entailment, contradiction, or not mentioned, with supporting evidence spans. We use these human annotations directly and do not create any contradictions ourselves. We selected the 52 NDAs with the most contradiction labels, since random sampling would mostly yield entailed or not-mentioned combinations. Across these documents, 213 of the 884 combinations are labeled contradictions with 409 evidence spans; each span forms one ground-truth pair, with the hypothesis as clause 1 and the span text as clause 2. This hypothesis-to-clause setup differs from the policy-to-policy comparison in GOV and USF, and only the 12 hypotheses that ever receive a contradiction label serve as queries, since the other five cannot produce true positives. Under our taxonomy, the NDA ground truth concentrates in specificity (202 pairs) and permission/prohibition (201), with six pairs elsewhere. The framework treats hypotheses as ordinary clauses during retrieval, so the architecture needs no changes. The following pair illustrates the setup.

\begin{quote}
\small
\textbf{Hypothesis}: ``Receiving Party may \textbf{retain some} Confidential Information even after the return or destruction of
Confidential Information.'' \\
\textbf{NDA clause}: ``[Confidential Information shall] remain the property of the Disclosing Party, and shall be \textbf{returned} by the Receiving Party \ldots\ (along with \textbf{all copies} thereof) promptly upon its receipt of a request to do so.''
\end{quote}

\noindent
The clause demands unconditional return of all copies on request, which cannot be reconciled with the retention right in the hypothesis. The framework embeds each hypothesis as a query clause, retrieves candidate NDA clauses in Stage~1, and confirms or rejects each (hypothesis, clause) pair in Stage~2. A detection counts as a true positive if it matches one of the 409 annotated pairs, using the same overlap-based matching as GOV and USF.

\begin{table}[t]
\centering
\small
\caption{Contradiction taxonomy with seven types used in ground-truth
annotation.}\label{tab:taxonomy}
\begin{tabular}{@{}lp{\dimexpr\columnwidth-3.4cm\relax}@{}}
\toprule
\textbf{Type} & \textbf{Definition} \\
\midrule
Negation          & Direct negation or modal conflict (``must'' vs.\ ``must not'') \\
Numerical         & Different numbers for the same metric and population \\
Temporal          & Incompatible deadlines, durations, or effective dates \\
Authority         & Different roles assigned sole/final authority over the same action \\
Process           & Different procedures prescribed for the same decision or action \\
Specificity       & A specific provision contradicts (not merely narrows) a general rule \\
Permission/Prohibition & One clause permits what another prohibits \\
\bottomrule
\end{tabular}
\end{table}

\subsection{Evaluation Metrics \& Models}\label{sec:metrics}

We report precision ($P$), recall ($R$), and F1 score. For retrieval evaluation, we use Recall at $K$ ($R@K$), the fraction of ground-truth contradictions for which at least one correct clause appears in the top-$K$ retrieved results, and Mean Reciprocal Rank~\citep[$\text{MRR} = \frac{1}{|Q|}\sum_{i=1}^{|Q|}\frac{1}{\text{rank}_i}$;][]{voorhees1999trec}, the average reciprocal rank at which the first relevant result appears for each query. A detection is a true positive if both of its clause texts match a ground-truth entry. Matching uses the overlap coefficient, the ratio of shared lowercased word tokens to the smaller token set, with a threshold of $\geq 0.35$. Since contradictions are unordered pairs, both clause assignments are tested, and the higher-scoring alignment is kept. Detections are assigned to ground-truth entries by optimal one-to-one matching with the Hungarian algorithm~\citep{kuhn1955hungarian}, which prevents mis-assignments among near-duplicate entries. Because ground-truth contradictions are injected into real policy documents, retrieval occasionally pairs an injected clause with an unrelated real clause instead of its intended partner. We treat an unmatched detection as such an injection artifact if either of its clauses overlaps a ground-truth clause at $\geq 0.70$, and artifacts are not counted as false positives. Both thresholds were set by examining matching behavior, 0.35 reliably separates true matches from topical overlap once contextual prefixes are stripped, and shifting the artifact threshold by $\pm 0.05$ changes F1 by at most 0.01. We assess statistical significance using paired bootstrap 95\% confidence interval \citep{berg2012empirical}, and exact McNemar's tests \citep{dror2018hitchhiker}.

Stage~2 uses four LLMs, GPT-5.1, Mistral Large 3.2, Phi-4, and Llama-4-Maverick, with smaller open-weight models (Qwen-2.5, Llama-3.1/3.2) for the skills analyses. Stage~1 generation (document summaries, contextual sentences, and HyDE hypotheses) uses GPT-5-mini. The GPT models run on Azure OpenAI, Mistral Large 3.2 and Llama-4-Maverick on Azure AI endpoints, and the small models locally via Ollama on three NVIDIA Tesla V100 GPUs. All calls use temperature~$= 0$ (greedy decoding).

\subsection{Baselines}\label{sec:baselines}

In the clause-only baseline, each candidate pair is sent to the LLM in isolation, one pair per request, without surrounding document context. The prompt states that Stage 1 has flagged these pairs as likely contradictions and that the verifier's job is to reject false positives. The verifier returns a JSON object with a verdict (confirmed or not a contradiction), a confidence level (high or low), and a reason. We also compare against a pretrained NLI model (DeBERTa-v3-base \citet{he2023debertav3}), which replaces the Stage 2 verifier on GOV and USF and scores every hypothesis against every clause on NDA. We aggregate by (hypothesis, document), flag pairs whose maximum contradiction probability exceeds a threshold, and sweep thresholds from 0.1 to 0.9.

Finally, we compare against direct prompting. Each policy document is sent to the LLM with instructions to identify all self-contradictions. Only self-contradictions are possible here because pairwise ones span documents the model never sees together. 

\section{\uppercase{Results}}\label{sec:results}

We report CLASH results in framework order. Stage~1 candidate generation (Section~\ref{sec:retrieval_ablations}), followed by Stage~2
verification (Section~\ref{sec:stage2_results}). Not all experiments use every corpus. The retrieval ablation experiments use the two HR corpora.
In these corpora, retrieval recall can be measured against clause-to-clause ground truth. The NDA corpus has 12 hypotheses that act as fixed queries and is used as a comparison corpus against the NLI models typically used in ContractNLI (Table~\ref{tab:stage2}). CPM appears only in the contradiction discovery analysis (Section~\ref{sec:organic}).

\subsection{Stage~1 Retrieval Results}\label{sec:retrieval_ablations}

\paragraph{Clause Representation}\label{sec:ctx_vs_raw}

To evaluate H2a, Table~\ref{tab:ctx_raw} compares retrieval recall across four clause representations on both policy corpora. Contextual representation achieves the highest recall. Adding explicit metadata tags (topic, population, role) to the contextual sentence does not improve recall and degrades MRR across both corpora, implying that the contextual sentence already captures the relevant semantic information. Broad unfiltered extraction performs poorly due to noise from non-policy sentences. The gains are largest on USF, where R@20 improves from 74\% (filtered raw) to 98\% (contextual). The broad unfiltered corpus is nearly unusable for USF (R@20 = 20\%), as the raw policy documents contain a lot of non-policy text, preambles, and revision histories that dilute the embedding space. 

\begin{table}[t]
\centering
\caption{Stage 1 retrieval ablation Clause representation comparison on both policy corpora.  All use Nomic Embed (768d). Contextual enrichment stays ahead at every retrieval depth $K$ on both corpora.}
\label{tab:ctx_raw}
\resizebox{\columnwidth}{!}{%
\begin{tabular}{@{}llccccc@{}}
\toprule
Corpus & Representation & R@5 & R@10 & R@20 & R@50 & MRR \\
\midrule
GOV & Broad unfiltered     & 51\% & 63\% & 76\% & 81\%  & .365 \\
GOV & Filtered raw         & 54\% & 69\% & 80\% & 87\%  & .385 \\
GOV & \textbf{Contextual} & \textbf{60\%} & \textbf{73\%} & \textbf{81\%} & \textbf{92\%} & \textbf{.444} \\
GOV & Ctx + metadata       & 57\% & 71\% & 80\% & 90\%  & .378 \\
\midrule
USF & Broad unfiltered     & 16\% & 18\% & 20\% & 22\% & .120 \\
USF & Filtered raw         & 66\% & 74\% & 74\% & 76\% & .531 \\
USF & \textbf{Contextual} & \textbf{94\%} & \textbf{96\%} & \textbf{98\%} & \textbf{100\%} & \textbf{.726} \\
USF & Ctx + metadata       & 88\% & 94\% & 96\% & 98\% & .698 \\
\bottomrule
\end{tabular}}
\end{table}

\begin{table}[t]
\centering
\small
\caption{Retrieval methods on both corpora, with the best value in each column shown in bold. The HyDE variants perform best on GOV, while all methods reach saturation on USF. We selected HyDE-Reranked, as it achieves the highest MRR on the harder corpus and maintains comparable performance on the saturated corpus.}
\label{tab:retrieval_methods}
\begin{tabular}{@{}lcccc@{}}
\toprule
& \multicolumn{2}{c}{GOV} & \multicolumn{2}{c}{USF} \\
\cmidrule(lr){2-3} \cmidrule(lr){4-5}
Method & R@20 & MRR & R@20 & MRR \\
\midrule
Bi-encoder    & 81\% & .444 & 98\%  & .726 \\
BM25          & 78\% & .406 & 98\%  & .712 \\
Hybrid RRF    & 82\% & .439 & \textbf{100\%} & \textbf{.754} \\
HyDE-only     & \textbf{92\%} & .467 & \textbf{100\%} & \textbf{.754} \\
HyDE-Reranked & 88\% & \textbf{.555} & \textbf{100\%} & .729 \\
\bottomrule
\end{tabular}
\end{table}

\paragraph{Choice of $K$.}
We select $K = 25$ for the policy corpora (GOV, USF) and $K = 20$ for NDA as the retrieval depth. Using $K = 25$ for a corpus of $M = 1{,}755$ clauses generates $1{,}755 \times 25 = 43{,}875$ directed pairs. After deduplication, this yields approximately 31,000 unique candidate pairs that the Stage~1 LLM must evaluate. If $K$ is set higher than 25, recall improves only slightly, but the number of pairs the LLM must process grows significantly. 

\paragraph{Retrieval Method Comparison}

Table~\ref{tab:retrieval_methods} compares retrieval methods on both corpora. Besides the bi-encoder baseline and HyDE variants, we evaluate Hybrid RRF, which fuses the bi-encoder and BM25 ranked lists via Reciprocal Rank Fusion ($k{=}60$)~\citep{cormack2009reciprocal}. 
On GOV, HyDE-Reranked achieves R@20 of 88\%, outperforming the bi-encoder’s 81\%. The largest improvements occur when the contradicting clause uses different vocabulary, such as specificity, numerical, and authority types. We select this approach over HyDE-only, despite HyDE-only's higher R@20 (92\%), because its higher MRR (0.555 vs. 0.467) places true contradictions in earlier batches with stronger candidates, making it easier for the Stage 1 screen to identify them. On USF, retrieval is nearly saturated. Hybrid RRF, HyDE-only, and HyDE-Reranked all reach 100\% R@20. Hybrid RRF and HyDE-only attain the highest MRR (0.754). BM25 performs better on USF (MRR 0.712) than on GOV (0.406), indicating lexical overlap carries more signal in the USF regulations. So, the best retrieval method depends on the corpora.

\paragraph{Stage~1 Candidate Generation Summary}\label{sec:stage1_results}

Table~\ref{tab:stage1} summarizes Stage~1 output across all three corpora. The LLM screen works as a high-recall filter. It retains 92 to 100\% of the ground-truth pairs but produces roughly 2 to 10 times as many false positives as true positives. This is intended since Stage 2 verification removes false positives.

\begin{table}[t]
\centering
\caption{Stage~1 candidate generation summary. False positives exclude injection artifacts (see Evaluation Metrics), so TP and FP need not sum to the candidate count.}
\label{tab:stage1}
\begin{tabular}{lcccc}
\toprule
Corpus & Candidates & TP & FP & Recall \\
\midrule
GOV   & 490     & 92  & 214     & .920 \\
USF   & 844     & 50  & 506     & 1.00 \\
NDA   & 1{,}084 & 395 & 689     & .988 \\
\bottomrule
\end{tabular}
\end{table}

\begin{table*}[!t]
\centering
\footnotesize
\caption{Stage~2 verification results. Context-aware verification reads each pair with its surrounding clauses and beats clause-only reading for every model. The NLI baseline performs significantly worse than all LLM verifiers.}
\label{tab:stage2}
\begin{tabular}{@{}llcccccl@{}}
\toprule
Model & Method & TP & FP & FN & P & R & F1 \\
\midrule
\multicolumn{8}{l}{\textit{GOV}} \\
\midrule
GPT-5.1       & Clause-only    & 81  & 29 & 19 & .736 & .810 & .771 \\
GPT-5.1       & Context-aware  & 76  & 5  & 24 & .938 & .760 & \textbf{.840} \\
Mistral-Large & Clause-only    & 77  & 48 & 23 & .616 & .770 & .684 \\
Mistral-Large & Context-aware  & 71  & 19 & 29 & .789 & .710 & \textbf{.747} \\
Llama-4-Mav.  & Clause-only    & 74  & 31 & 26 & .705 & .740 & .722 \\
Llama-4-Mav.  & Context-aware  & 69  & 14 & 31 & .831 & .690 & \textbf{.754} \\
Phi-4 (14B)   & Clause-only    & 92  & 214 & 8 & .301 & .920 & .453 \\
Phi-4 (14B)   & Context-aware  & 65  & 9  & 35 & .878 & .650 & \textbf{.747} \\
DeBERTa-v3-base & NLI & 62 & 41 & 38 & .602 & .620 & .611 \\
\midrule
\multicolumn{8}{l}{\textit{USF}} \\
\midrule
GPT-5.1       & Clause-only    & 43  & 29 & 7  & .597 & .860 & .705 \\
GPT-5.1       & Context-aware  & 43  & 6  & 7  & .878 & .860 & \textbf{.869} \\
Mistral-Large & Clause-only    & 40  & 37 & 10 & .519 & .800 & .630 \\
Mistral-Large & Context-aware  & 40  & 11 & 10 & .784 & .800 & \textbf{.792} \\
Llama-4-Mav.  & Clause-only    & 40  & 46 & 10 & .465 & .800 & .588 \\
Llama-4-Mav.  & Context-aware  & 41  & 44 & 9  & .482 & .820 & .607 \\
Phi-4 (14B)   & Clause-only    & 50  & 506 & 0 & .090  & 1.00  & .165 \\
Phi-4 (14B)   & Context-aware  & 32  & 13 & 18 & .711 & .640 & .674 \\
DeBERTa-v3-base & NLI & 38 & 74 & 12 & .339 & .760 & .469 \\

\midrule
\multicolumn{8}{l}{\textit{NDA}} \\
\midrule
GPT-5.1       & Hyp.-to-clause & 333 & 3   & 76  & .991 & .814 & .894 \\
Mistral-Large & Hyp.-to-clause & 396 & 78  & 13  & .835 & .968 & .897 \\
Llama-4-Mav.  & Hyp.-to-clause & 374 & 29  & 35  & .928 & .914 & \textbf{.921} \\

DeBERTa-v3-base       & NLI   & 212 & 664 & 197 & .242 & .518 & .330 \\
DeBERTa-v3-large      & NLI   & 210 & 636 & 199 & .248 & .513 & .335 \\
\bottomrule
\end{tabular}
\end{table*}

\begin{table*}[!t]
\centering
\footnotesize
\caption{Context window ablation on two models.}
\label{tab:ctx_window}
\begin{tabular}{l ccc ccc ccc ccc}
\toprule
 & \multicolumn{3}{c}{\textbf{GOV -- Mistral}} & \multicolumn{3}{c}{\textbf{GOV -- Llama-4-Mav.}} & \multicolumn{3}{c}{\textbf{USF -- Mistral}} & \multicolumn{3}{c}{\textbf{USF -- Llama-4-Mav.}} \\
\cmidrule(lr){2-4} \cmidrule(lr){5-7} \cmidrule(lr){8-10} \cmidrule(lr){11-13}
Window & P & R & F1 & P & R & F1 & P & R & F1 & P & R & F1 \\
\midrule
$\pm 1$  & .803 & .650 & .718 & .753 & \textbf{.760} & \textbf{.756} & .816 & .800 & .808 & .377 & .800 & .513 \\
$\pm 3$  & \textbf{.831} & .690 & \textbf{.754} & .758 & .750 & .754 & .789 & \textbf{.820} & .804 & .472 & \textbf{.840} & .604 \\
$\pm 5$  & .789 & .710 & .747 & \textbf{.831} & .690 & .754 & .784 & .800 & .792 & .482 & .820 & .607 \\
$\pm 7$  & .786 & .660 & .717 & .810 & .680 & .739 & \textbf{.820} & \textbf{.820} & \textbf{.820} & .541 & .800 & .645 \\
$\pm 10$ & .783 & \textbf{.720} & .750 & .785 & .620 & .693 & .804 & .740 & .771 & \textbf{.722} & .780 & \textbf{.750} \\
\bottomrule
\end{tabular}
\end{table*}

\subsection{Stage~2: Verification Results}\label{sec:stage2_results}

Stage~2 applies context-aware verification to each Stage~1 candidate. To test H2b, we compare it with clause-only verification on GOV and USF (Table~\ref{tab:stage2}). Consistent with H2b, context-aware verification improves precision while keeping comparable recall ($p<0.05$). On both HR corpora, all models reduce false positives significantly at the cost of a few recall points. The surrounding clauses help the verifier recognize the most common false positive pattern (general versus specific exceptions), and complementary provisions that clause-only verification mistakes for contradictions. With the $\pm 5$ window, the exception is visible, and adjacent text clarifies that different provisions apply to different categories. Most misses are pairwise. The verifiers find 82--94\% of self-contradictions on GOV but only 55--59\% of pairwise ones. The gap is smaller on USF, with 85--89\% versus 74--83\%. Recall on subtle pairs is 12--22 points lower than on obvious pairs on GOV and at most 11 points lower on USF. The 29 distractor documents test false positive resistance on GOV. Because unrelated policies use similar administrative language, distractor pairs make up 58\% of the original retrieval candidates, 27\% after Stage~1 screening, and less than 9\% of confirmed cases. USF carries out the same test naturally. A rule applicable to its faculty (twelve regulations) cannot contradict a rule applicable to its staff (thirteen), yet the proportion of cross-group pairs increases from 15.4\% of the 21,715 original candidates to 19.8\% of the 844 that pass Stage~1. The context-aware verifier resolves which population each rule covers, and almost none of its false positives cross this division. With ten parallel requests, GPT-5.1 verifies 490 GOV candidates in 73 seconds and 844 USF candidates in 129 seconds. Mistral-Large takes 4 to 9 minutes per corpus and Llama-4-Maverick 21 to 52 minutes, mostly due to slower serving endpoints. Stage~1 keeps 92 of 100 GOV pairs and all 50 USF pairs, so most GPT-5.1 misses (16 of 24 on GOV, all 7 on USF) are verifier rejections. In most of these, the verifier decides the two clauses govern different situations. 

\paragraph{Context Window Size}\label{sec:ablation_ctx_window}

The context-aware verifier retrieves $\pm N$ surrounding clauses from each source document. To select a proper window size, we sweep $N \in \{1, 3, 5, 7, 10\}$ on both policy corpora with Mistral-Large and Llama-4-Maverick. Table~\ref{tab:ctx_window} reports precision, recall, and F1 for each configuration. A wider window generally improves precision but may reduce recall on GOV policies. The USF regulations benefit from more surrounding text. No single window wins everywhere, but $\pm 5$ performs well across models and corpora; therefore, we use $\pm 5$ as the default in all main results (Table~\ref{tab:stage2}).

\subsection{Baseline Comparisons}\label{sec:baseline_results}

To answer RQ1, we compare the decomposition against both formulations. The NLI baseline follows every LLM verifier on every corpus. DeBERTa-v3-base reaches F1 = .611 on GOV, .469 on USF, and .330 on NDA, against the framework's .840, .869, and .921 (Table~\ref{tab:stage2}). On NDA, we also test a larger variant, DeBERTa-v3-large, trained on a broader mix of NLI datasets, and it improves the result by only half a point. It demonstrates that an NLI-based model isn’t suitable for contradiction detection in policy language settings. Also, an NLI model judges each sentence pair in isolation, so the contradiction types that depend on document context suffer most, with specificity at 39\% recall and authority and temporal types almost always missed.

For whole-document prompting, each policy document is sent in full to the LLM with instructions to identify all self-contradictions, since pairwise contradictions span documents the model never sees together. To keep the comparison fair, both systems are scored only on the self-contradiction subset of the ground truth (51 GOV pairs, 27 USF pairs), and the framework's detections are restricted to same-document pairs. Direct prompting keeps recall above 90\%, but precision stays below 14\%, giving F1 between .201 and .232 for Mistral-Large and Llama-4-Maverick, while the framework scored on the same subset reaches .640 to .809, a gap of 41--61 points. All framework-baseline differences in this section are statistically significant ($p<0.05$). Direct prompting also produces 7--8$\times$ more false positives than true positives, which makes its output impractical for human review. 

\begin{table}[htb!]
\small
\centering
\begin{tabular}{@{}lcc@{}}
\toprule
Method & GOV F1 & USF F1 \\
\midrule
Context-aware verifier (no tools) & \textbf{.840} & \textbf{.869} \\
Confidence cascade & .837 & .672 \\
Tool-augmented single agent & .690 & .731 \\
\bottomrule
\end{tabular}
\caption{Best F1 per agentic design on the standard Stage~1 candidates. No design outperforms the single-pass context-aware verifier on either corpus.}
\label{tab:agentic}
\end{table}

\begin{table}[!t]
\centering
\footnotesize
\setlength{\tabcolsep}{3pt}
\caption{Skills based verification, context-aware baseline vs. +Skills. Type-specific skills significantly increase recall on small open-weight models.}
\label{tab:skills}
\begin{tabular}{@{}l ccc ccc@{}}
\toprule
& \multicolumn{3}{c}{\textbf{Baseline (ctx-aware)}} & \multicolumn{3}{c}{\textbf{+Skills}} \\
\cmidrule(lr){2-4} \cmidrule(lr){5-7}
Model & P & R & F1 & P & R & F1 \\
\midrule
\multicolumn{7}{l}{\textit{Frontier models, GOV}} \\
\midrule
GPT-5.1        & .938 & .760 & \textbf{.840} & .925 & .740 & .822 \\
Mistral-Large  & .789 & .710 & .747 & .738 & .760 & \textbf{.749} \\
\midrule
\multicolumn{7}{l}{\textit{Frontier models, USF}} \\
\midrule
GPT-5.1        & .878 & .860 & \textbf{.869} & .741 & .860 & .796 \\
Mistral-Large  & .784 & .800 & \textbf{.792} & .677 & .880 & .765 \\
\midrule
\multicolumn{7}{l}{\textit{Small open-weight models, GOV}} \\
\midrule
Llama-3.2 (3B) & .545 & .060 & .108 & .433 & .680 & \textbf{.529} \\
Qwen-2.5 (14B) & .877 & .570 & .691 & .786 & .810 & \textbf{.798} \\
Phi-4 (14B)    & .878 & .650 & .747 & .708 & .800 & \textbf{.751} \\
\midrule
\multicolumn{7}{l}{\textit{Small open-weight models, USF}} \\
\midrule
Llama-3.2 (3B) & .312 & .100 & .152 & .143 & .860 & \textbf{.246} \\
Qwen-2.5 (14B) & .576 & .760 & \textbf{.655} & .381 & .900 & .536 \\
Phi-4 (14B)    & .711 & .640 & \textbf{.674} & .404 & .840 & .545 \\
\bottomrule
\end{tabular}
\end{table}

\subsection{Tool- and Skills-Based Verification}\label{sec:skills}

The last part of RQ2 examines whether using agentic tools and type-specific skills helps with verification. We examine two agentic tool designs and compare them with the single-pass context-aware verifier. One is a tool-augmented single agent that can use retrieval and search tools while judging a pair. The other is a confidence cascade that rechecks pairs only when the context-aware verifier is not confident. For all runs, we use the models listed in Section~\ref{sec:metrics} and the same Stage~1 candidates. Table~\ref{tab:agentic} shows the best result for each design. Additionally, we assess type-specific specialist sub-agents, multi-agent debate, and multi-round tool calling, but all stay between .62 and .69 F1 on both corpora. Contrary to H2c, none of the agentic designs outperforms the context-aware verifier. There are three main reasons for this. First, task and tools mismatch: verification is a bounded task whose evidence fits in a single prompt, so the $\pm 5$ pre-loaded context is enough. Adding retrieval adds noise. Second, an evidence paradox arises. Tools retrieving broader context give the verifier unrelated information to explain away the conflict. Third, complexity overhead: each extra LLM call can fail in its own way, like hallucinated tool names or parsing errors. The cascade results match our context-aware verifier on GOV, keep the single-pass approach, and escalate only low-confidence pairs. So agentic designs might still help when escalating certain cases is necessary. 

Recent work on skill-augmented LLM agents shows that injecting compact, task-specific instructions (``skills'') into prompts can improve performance, with 2--3 focused skills outperforming larger libraries~\citep{li2026skillsbench}. Following this idea, we extend the context-aware verifier with type-specific skills, 20--25-line instruction blocks added to the verifier prompt that give accept and reject patterns for three families of contradiction types. Value conflict covers numerical, temporal, and specificity types, permission conflict covers permission/prohibition and negation, and authority/procedure conflict covers authority and process. Each skill is activated via type-gated routing based on Stage~1's contradiction-type label. Table~\ref{tab:skills} shows that for frontier models, skills do not help because the base verification prompt already provides everything the skills add. For small models, the gains on GOV are large. Llama-3.2's F1 jumps from .108 to .529 and Qwen-2.5 from .691 to .798. On both corpora, skills raise recall for all three models, but on USF precision drops so far that F1 falls below the baseline for the two 14B models. Recall improvements across both corpora are statistically significant (all $p<0.05$). So, skills help when the base model is too weak to follow the verification prompt on its own, and hurt once the model already follows it.

\subsection{Contradiction Discovery in Real Policies}\label{sec:organic}

We address RQ3 by reviewing each Stage~2 false positive, since some of these pairs could be real conflicts in the original documents. Four LLMs vote on every pair, and we manually verify the shortlisted candidates against the source documents. The consensus rejected all 184 GOV pairs, confirmed three previously unknown contradictions on USF, and reduced 30 CPM candidates to two confirmed conflicts. Consistent with H3, the framework found five real contradictions across two organizations and produced zero false discoveries on the clean corpus.

\noindent\textbf{Form naming (USF).} Policy 0-108 requires discrimination reports ``using the Equal
  Opportunity Report Form,'' while Policy 0-007 requires ``the Equal Opportunity Compliance Inquiry Form''
  for the same report.

\noindent\textbf{Grievance rights (USF).} Reg.~10.206 states that an employee without permanent status
  ``may be terminated at any time without the right to grieve such action,'' while Reg.~10.213 allows
  non-permanent employees to grieve the wrongful application of a regulation.

\noindent\textbf{Pay status (USF).} Reg.~10.202 requires that an employee ``must be in pay status to
  receive a pay increase of any kind,'' yet guarantees employees returning from unpaid leave the ``mandatory pay increases granted during the period of unpaid leave.''

\noindent\textbf{Bonus eligibility (CPM).} Volume~1404 requires a rating of Fully Successful for a performance bonus, while Volume~1405 requires Achieved Expectations with all performance elements met.

\noindent\textbf{Hiring priority (CPM).} Volume~1471 gives Reemployment Priority List placement precedence over military spouse preference, while Volume~315 requires selecting the military spouse when among the best qualified candidates.

\section{\uppercase{Limitations and Future Work}}
CLASH has a few limitations. Our ground truth consists of contradictions injected into real documents, the dominant evaluation practice in this field~\citep{schumann2024literature}, and the five actual contradictions found in USF and CPM show that the framework transfers beyond injected cases. Also, all corpora are in English and end-to-end recall is bounded by Stage~1, since a pair that retrieval misses can never reach verification. The framework also requires clause extraction as a pre-processing step, and segmentation errors propagate into both stages. Future work will improve recall on pairwise contradictions and extend the framework to non-English policies. 

\section{\uppercase{Conclusion}}\label{sec:conclusion}

We introduced CLASH, a two-stage framework that first generates candidate pairs with a focus on recall, then verifies each pair with document context. RQ1 asked whether this division outperforms NLI-based classification and whole-document prompting on multi-document policy corpora. The results confirm H1: CLASH outperforms the NLI baselines by 23--59 F1 points and whole-document prompting by 41--61 points across three corpora, and the baselines fail in complementary ways. NLI cannot judge conflicts that need document context, and direct prompting buries real conflicts under false positives. RQ2 asked whether context improves both stages and whether agentic tool use improves verification. We found context helped at both stages, and tool use did not. As H2a and H2b predicted, contextual clause representation raised retrieval recall from 74\% to 98\%, and surrounding clauses at verification cut false positives significantly. H2c is the one expectation the results overturned. None of the agentic configurations we tested beats the single-pass context-aware verifier, and type-specific skills help only small open-weight models that cannot follow the base prompt on their own. A verifier needs the right evidence in its prompt more than it needs machinery around it. RQ3 asked whether the framework can reveal genuine, previously unknown contradictions in policy corpora. H3 expected that some detections matching no ground truth entry would be real conflicts, and manual review against the source documents confirmed that. We found five contradictions that already existed in the original policies, three in USF, two in CPM. Organizations that revise policies individually often accumulate such conflicts, which previously required manual review to identify. CLASH shows that these issues can be addressed automatically, with sufficient precision for human verification.

\bibliographystyle{apalike}
{\small
\bibliography{aaai2026}}

\begin{thebibliography}{}

\bibitem[Abualhaija et~al., 2024]{abualhaija2024ai}
Abualhaija, S., Ceci, M., Sannier, N., Bianculli, D., Briand, L.~C., Zetzsche,
  D., and Bodellini, M. (2024).
\newblock Ai-enabled regulatory change analysis of legal requirements.
\newblock In {\em 2024 IEEE 32nd International Requirements Engineering
  Conference (RE)}, pages 5--17. IEEE.

\bibitem[Aires et~al., 2017]{aires2017norm}
Aires, J.~P., Pinheiro, D., Lima, V. S.~d., and Meneguzzi, F. (2017).
\newblock Norm conflict identification in contracts.
\newblock {\em Artificial Intelligence and Law}, 25(4):397--428.

\bibitem[Anthropic, 2024]{anthropic2024contextual}
Anthropic (2024).
\newblock Introducing contextual retrieval.
\newblock \url{https://www.anthropic.com/engineering/contextual-retrieval}.
\newblock Accessed: 2026-07-21.

\bibitem[Berg-Kirkpatrick et~al., 2012]{berg2012empirical}
Berg-Kirkpatrick, T., Burkett, D., and Klein, D. (2012).
\newblock An empirical investigation of statistical significance in nlp.
\newblock In {\em Proceedings of EMNLP-CoNLL 2012}, pages 995--1005.

\bibitem[Bowman et~al., 2015]{bowman2015snli}
Bowman, S.~R., Angeli, G., Potts, C., and Manning, C.~D. (2015).
\newblock A large annotated corpus for learning natural language inference.
\newblock In {\em Proceedings of EMNLP 2015}, pages 632--642.

\bibitem[Choudhury et~al., 2026]{choudhury2026better}
Choudhury, M.~R., Chandramouli, A., Anand, M., and Gupta, V. (2026).
\newblock Better call clause: A discrepancy benchmark for auditing llms legal
  reasoning capabilities.
\newblock In {\em Findings of the Association for Computational Linguistics:
  EACL 2026}, pages 5776--5818.

\bibitem[Conti et~al., 2025]{conti2025context}
Conti, M., Faysse, M., Viaud, G., Bosselut, A., Hudelot, C., and Colombo, P.
  (2025).
\newblock Context is gold to find the gold passage: Evaluating and training
  contextual document embeddings.
\newblock In {\em Proceedings of EMNLP 2025}, pages 22605--22619.

\bibitem[Cormack et~al., 2009]{cormack2009reciprocal}
Cormack, G.~V., Clarke, C.~L., and Buettcher, S. (2009).
\newblock Reciprocal rank fusion outperforms condorcet and individual rank
  learning methods.
\newblock In {\em Proceedings of SIGIR 2009}, pages 758--759.

\bibitem[Douze et~al., 2025]{douze2025faisslibrary}
Douze, M., Guzhva, A., Deng, C., Johnson, J., Szilvasy, G., Mazaré, P.-E.,
  Lomeli, M., Hosseini, L., and Jégou, H. (2025).
\newblock The faiss library.

\bibitem[Dror et~al., 2018]{dror2018hitchhiker}
Dror, R., Baumer, G., Shlomov, S., and Reichart, R. (2018).
\newblock The hitchhiker’s guide to testing statistical significance in
  natural language processing.
\newblock In {\em Proceedings of ACL 2018}, pages 1383--1392.

\bibitem[Elenga~G\"{a}rtner and G\"{o}hlich, 2024]{gaertner2024alice}
Elenga~G\"{a}rtner, A. and G\"{o}hlich, D. (2024).
\newblock Automated requirement contradiction detection through formal logic
  and {LLMs}.
\newblock {\em Automated Software Engineering}, 31(49).

\bibitem[Gao et~al., 2023]{gao2023hyde}
Gao, L., Ma, X., Lin, J., and Callan, J. (2023).
\newblock Precise zero-shot dense retrieval without relevance labels.
\newblock In {\em Proceedings of ACL 2023}, pages 1762--1777.

\bibitem[Gokul et~al., 2025]{gokul2025ragcontradiction}
Gokul, V., Tenneti, S., and Nakkiran, A. (2025).
\newblock Contradiction detection in {RAG} systems: Evaluating {LLMs} as
  context validators for improved information consistency.
\newblock {\em arXiv preprint arXiv:2504.00180}.

\bibitem[He et~al., 2023]{he2023debertav3}
He, P., Gao, J., and Chen, W. (2023).
\newblock {DeBERTaV3}: Improving {DeBERTa} using {ELECTRA}-style pre-training
  with gradient-disentangled embedding sharing.
\newblock In {\em Proceedings of ICLR 2023}.

\bibitem[He et~al., 2026]{he2026foundations}
He, P., Kirschbaum, E., and Kasiviswanathan, S. (2026).
\newblock Foundations of global consistency checking with noisy llm oracles.
\newblock {\em arXiv preprint arXiv:2601.13600}.

\bibitem[Huang et~al., 2024]{huang2024neuralconflict}
Huang, S., Sun, J., and Li, R. (2024).
\newblock Neuralconflict: Using neural networks to identify norm conflicts in
  normative documents.
\newblock {\em Expert Systems}, 41(6):e13035.

\bibitem[Jalocha and Michelsen, 2026]{jalocha2026crawler}
Jalocha, M. and Michelsen, W. (2026).
\newblock {CRAwLeR}---cross-reference aware legal retrieval.
\newblock {\em arXiv preprint arXiv:2606.21676}.

\bibitem[Jeong et~al., 2026]{jeong2026toolmad}
Jeong, S., Choi, Y., Kim, J., and Jang, B. (2026).
\newblock {Tool-MAD}: A multi-agent debate framework for fact verification with
  diverse tool augmentation and adaptive retrieval.
\newblock {\em arXiv preprint arXiv:2601.04742}.

\bibitem[Jiang et~al., 2026]{jiang2026sok}
Jiang, Y., Li, D., Deng, H., Ma, B., Wang, X., Wang, Q., and Yu, G. (2026).
\newblock Sok: Agentic skills--beyond tool use in llm agents.
\newblock {\em arXiv preprint arXiv:2602.20867}.

\bibitem[Koreeda and Manning, 2021]{koreeda2021contractnli}
Koreeda, Y. and Manning, C.~D. (2021).
\newblock {ContractNLI}: A dataset for document-level natural language
  inference for contracts.
\newblock In {\em Findings of the Association for Computational Linguistics:
  EMNLP 2021}, pages 1907--1919.

\bibitem[Kozlova et~al., 2026]{kozlova2026contradiction}
Kozlova, E., Petrova, M., and Baiuk, A. (2026).
\newblock Contradiction detection in administrative offense rulings: A case
  study analysis.
\newblock In {\em Proceedings of the International Conference ``Dialogue''},
  volume 2026.

\bibitem[Kuhn, 1955]{kuhn1955hungarian}
Kuhn, H.~W. (1955).
\newblock The {Hungarian} method for the assignment problem.
\newblock {\em Naval Research Logistics Quarterly}, 2(1-2):83--97.

\bibitem[Kurfal{\i}, 2025]{kurfali2025conflicting}
Kurfal{\i}, M. (2025).
\newblock Conflicting needles in a haystack: How llms behave when faced with
  contradictory information.
\newblock In {\em Proceedings of EMNLP 2025}, pages 34349--34364.

\bibitem[Li et~al., 2024]{li2024contradoc}
Li, J., Raheja, V., and Kumar, D. (2024).
\newblock Contradoc: Understanding self-contradictions in documents with large
  language models.
\newblock In {\em Proceedings of NAACL-HLT 2024}, pages 6509--6523.

\bibitem[Li et~al., 2026]{li2026skillsbench}
Li, X., Liu, Y., Chen, W., You, B., Di, Z., He, Y., Zheng, S., Choe, K.~W.,
  Sun, J., et~al. (2026).
\newblock Skillsbench: Benchmarking how well agent skills work across diverse
  tasks.

\bibitem[Liu et~al., 2024]{liu2024lost}
Liu, N.~F., Lin, K., Hewitt, J., Paranjape, A., Bevilacqua, M., Petroni, F.,
  and Liang, P. (2024).
\newblock Lost in the middle: How language models use long contexts.
\newblock {\em Transactions of the Association for Computational Linguistics},
  12:157--173.

\bibitem[Ma et~al., 2026]{ma2026gkmad}
Ma, X., Rao, G., Xu, L., Wang, X., Fan, Z., and Zhang, Z. (2026).
\newblock Guided and knowledgeable multi-agent debate for fact verification.
\newblock {\em Expert Systems With Applications}, 299:130103.

\bibitem[MacCartney, 2009]{maccartney2009natural}
MacCartney, B. (2009).
\newblock {\em Natural language inference}.
\newblock Stanford University.

\bibitem[Mantravadi et~al., 2025]{mantravadi2025legalwiz}
Mantravadi, A., Dalmia, S., Mukherji, A., Dave, N., Mittal, A., and Pospelova,
  O. (2025).
\newblock {LegalWiz}: A multi-agent generation framework for contradiction
  detection in legal documents.
\newblock In {\em NeurIPS 2025 Workshop on Generative and Protective AI for
  Content Creation}.
\newblock arXiv:2510.03418.

\bibitem[Nussbaum et~al., 2024]{nussbaum2024nomic}
Nussbaum, Z., Morris, J.~X., Duderstadt, B., and Mulyar, A. (2024).
\newblock Nomic embed: Training a reproducible long context text embedder.
\newblock {\em arXiv preprint arXiv:2402.01613}.

\bibitem[OpenAI, 2025]{openai_gpt5_mini}
OpenAI (2025).
\newblock Gpt-5 mini model.
\newblock OpenAI Model Card / API Documentation.
\newblock \url{https://platform.openai.com/docs/models/gpt-5-mini}.

\bibitem[Sai et~al., 2023]{sai2023detecting}
Sai, C., Winter, K., Fernanda, E., and Rinderle-Ma, S. (2023).
\newblock Detecting deviations between external and internal regulatory
  requirements for improved process compliance assessment.
\newblock In {\em International Conference on Advanced Information Systems
  Engineering}, pages 401--416. Springer.

\bibitem[Schumann, 2025]{schumann2025contradictions}
Schumann, G. (2025).
\newblock Contradictions and inconsistencies in regulatory documents--a
  qualitative assessment from practice.
\newblock In {\em 2025 15th International Conference on Advanced Computer
  Information Technologies (ACIT)}, pages 273--282. IEEE.

\bibitem[Schumann and Marx~G\'{o}mez, 2024]{schumann2024literature}
Schumann, G. and Marx~G\'{o}mez, J. (2024).
\newblock Detection of conflicts, contradictions and inconsistencies in
  regulatory documents: A literature review.
\newblock In {\em 2024 Fifth International Conference on Intelligent Data
  Science Technologies and Applications (IDSTA)}, pages 81--88. IEEE.

\bibitem[Schumann et~al., 2026]{schumann2026detection}
Schumann, G., Marx~G{\'o}mez, J., and Pargmann, H. (2026).
\newblock Detection of contradictions and inconsistencies in regulatory
  documents using prompt-engineering.
\newblock In {\em Proceedings of HICSS 2026}.

\bibitem[Voorhees, 1999]{voorhees1999trec}
Voorhees, E.~M. (1999).
\newblock The {TREC-8} question answering track report.
\newblock In {\em Proceedings of the Eighth Text REtrieval Conference
  (TREC-8)}.

\bibitem[Williams et~al., 2018]{williams2018multinli}
Williams, A., Nangia, N., and Bowman, S.~R. (2018).
\newblock A broad-coverage challenge corpus for sentence understanding through
  inference.
\newblock In {\em Proceedings of NAACL-HLT 2018}, pages 1112--1122.

\end{thebibliography}

\end{document}